\pdfoutput=1

\documentclass[11pt]{article}

\usepackage[final]{acl}

\usepackage{times}
\usepackage{latexsym}
\usepackage{comment}
\usepackage{todonotes}
\usepackage[T1]{fontenc}

\usepackage[utf8]{inputenc}

\usepackage{microtype}

\usepackage{inconsolata}

\usepackage{graphicx}
\usepackage{coptic}
\usepackage{stackengine}
\usepackage{enumitem}
\usepackage{multirow}

\title{Lacuna Language Learning: Leveraging RNNs for Ranked Text Completion in Digitized Coptic Manuscripts}

\author{
  \textbf{Lauren Levine\textsuperscript{1}},
  \textbf{Cindy Tung Li\textsuperscript{1}}, \\
  \textbf{Lydia Bremer-McCollum\textsuperscript{2}},
  \textbf{Nicholas Wagner\textsuperscript{3}},
 \textbf{Amir Zeldes\textsuperscript{1}}
 \\
  \textsuperscript{1}Georgetown University,
  \textsuperscript{2}University of Oklahoma,
  \textsuperscript{3}Duke University
\\
    \texttt{\{lel76, cl1579, amir.zeldes\}@georgetown.edu} \\
    \texttt{lcbm@ou.edu}, \texttt{nicholas.wagner@duke.edu}
}

\begin{document}
\maketitle
\begin{abstract}

Ancient manuscripts are frequently damaged, containing gaps in the text known as lacunae. In this paper, we present a bidirectional RNN model for character prediction of Coptic characters in manuscript lacunae. Our best model performs with 72\% accuracy on single character reconstruction, but falls to 37\% when reconstructing lacunae of various lengths. While not suitable for definitive manuscript reconstruction,  we argue that our RNN model can help scholars rank the likelihood of textual reconstructions. As evidence, we use our RNN model to rank reconstructions in two early Coptic manuscripts. Our investigation shows that neural models can augment traditional methods of textual restoration, providing scholars with an additional tool to assess lacunae in Coptic manuscripts.

\end{abstract}
\section{Introduction}

Ancient manuscripts are an invaluable resource for linguists and historians, offering insights into the cultures and languages of the ancient world. Unfortunately, these manuscripts are often damaged, with sections of text missing, known as lacunae. In recent years, neural models have made significant advances in various areas of linguistic research. Nevertheless, attempts to apply neural methods to manuscript reconstruction have been limited, and none have specifically targeted Coptic (see Section \ref{sec:background_reconstruction}).

In this paper, we explore the potential for neural language models to be utilized in the reconstruction of Coptic manuscripts. Leveraging a bidirectional RNN language model trained for Coptic character prediction, we explore how the model can be integrated into the workflow of scholars attempting textual reconstruction. We consider the ability of the model to predict the missing characters of lacunae directly, as well as to provide rankings for the likelihood of reconstruction candidates already under consideration. We show that scholars can use judgments from neural models as additional quantitative evidence, in conjunction with more traditional qualitative methods, to work towards manuscript reconstruction.

\section{Background and Related Work}
\subsection{Coptic}

Coptic belongs to the Afro-Asiatic language family and is the latest stage of the Egyptian language, the longest continuously attested language on Earth. Coptic utilizes the 24 glyphs of the Greek alphabet and adds additional Demotic (Egyptian) glyphs (a minimum of 6 depending on dialect) to represent sounds not found in Greek. In late antiquity, more than a dozen regional dialects were spoken and written \cite{Layton2011}. Owing to these dialect variations, the use of superlinear strokes and other diacritical marks, and irregular orthography of Greek loan words, Coptic provides a highly complex dataset.

Coptic manuscripts preserve the diverse textual tradition of late-antique and medieval Egypt. Inscribed on papyrus and other perishable media, many Coptic manuscripts contain small gaps or holes (lacunae), which often cannot be restored on the basis of other extant manuscripts. Scholars use qualitative methods to restore lacunae, chiefly through study of the manuscript’s context and (con)textual parallels. Occasionally, appeal is made to traditional canons of textual criticism, but here too the scholar's own judgment guides the restoration \cite{Wasserman2013}. Initial testing has shown that human methods of textual restoration have a high error rate at both the word level and the character level \cite[711--712]{SommerschieldEtAl2023}.

\subsection{Manuscript Reconstruction}
\label{sec:background_reconstruction}

Following early attempts using n-gram models to approach the Indus Valley script \cite{RaoYadavVahiaEtAl2009}, most previous work on reconstructing lacunae in manuscripts, as well as in epigraphic data, has focused on Greek and Latin \cite{NovokhatkoMaier2022,Matsumoto2022}. Early projects included eAQUA \cite{Schubert2011}, which pioneered proposing automatic reconstructions of lacunae based on statistical methods from larger datasets  (in the context of ancient languages). More recently, studies using neural methods for the reconstruction of Greek \cite{assael-etal-2019-restoring} and Latin \cite{BrunelloCLMS023} have appeared, with papers in the last three years specifically proposing to leverage transformer based language model architectures for both born-digital and (OCRed) handwritten inputs in a range of languages \cite{vogler-etal-2022-lacuna}.

We are not aware of previous papers applying language models to the reconstruction of Coptic texts, though a recent Web page prepared by the CoptOT project\footnote{\url{https://coptot.manuscriptroom.com/manuscript-speculation-tool}} provides a `Manuscript Speculation Tool' which helps in laying out missing letters on predefined digitized manuscript spaces. However, in the tool's operating scenario, a base text to be laid out is known (e.g.~a chapter of the Bible), and the question is how many letters of each verse might fit into each missing line or part of a line. To our knowledge, this paper is the first attempt to leverage language modeling for lacuna reconstruction in Coptic.

\subsection{Masked Language Models}
In 2019, \citeauthor{devlin2019bert} introduced BERT, a foundational masked language model (MLM), where random tokens in the input were masked, and the model was trained to predict the masked token based on the context. For 15\% of the tokens in training, each one is replaced with either [MASK], a random token, or the original token, without change. Masking mimics gaps and teaches the model to fill in missing segments of strings, which makes the MLM approach highly applicable to our lacuna reconstruction task. 

In the same paper, Devlin et al. found that a model with only left to right context performed worse than a bidirectional masked language model, which is able to use context from before and after the masked token. They advocate for a bidirectional model that can use left and right context at every layer over concatenating a left to right model and a right to left model, as proposed earlier in ELMo \cite{peters-etal-2018-deep}. As we are  framing our lacuna reconstruction task as a prediction of masked characters, parallel to the masked token prediction done by models such as BERT, this finding regarding bidirectionality leads us to adopt a bidirectional strategy for our model as well.

As the masked language model strategy was popularized with transformer based models such as BERT, there is not much existing work regarding the implementation of masked language models with an RNN-based architecture. However, in scenarios with relatively small quantities of data and limited long distance dependencies, it can still be preferable to use an RNN-based architecture over a transformer-based architecture \cite{mishra2021}. Considering that we have almost 1.22 million tokens of Coptic data, and we are looking to fill in character gaps at the sentence level, we consider our Coptic lacuna prediction task to be one such scenario, and we opt to use an RNN based architecture in our model. 

While we have done some preliminary prototyping with transformer based architectures, such as ELECTRA \cite{clark2020electra}, so far our experiments with RNN-based architectures have made the most progress. As such, we present those finding here. However, we still believe it would be worthwhile to return to the exploration of various transformer architectures in future work.

\section{Data}
\label{data}

For training and testing the model, we leverage the data from the Coptic SCRIPTORIUM Corpora \cite{schroeder_raiders_2016}. This project compiles text from a variety of manuscript sources and totals almost 1.22 million tokens of Sahidic Coptic. The Coptic SCRIPTORIUM project is an ongoing effort to create an open online database and tool set for digital research in Coptic. This effort includes creating normalized, machine readable versions of Coptic manuscripts with a variety of linguistic annotations created using the online, version controlled GitDox annotation tools \cite{ZhangZeldes2017}. The full data set is publicly available on GitHub\footnote{\url{https://github.com/CopticScriptorium/corpora}} in various machine readable formats, and the corpora are searchable via an online query interface.\footnote{\url{https://annis.copticscriptorium.org/annis/scriptorium}}

The digitized manuscripts have a normalized version (with regard to spelling, etc.) of the text as well as a version representing the original text. We leverage the original text version, annotated as \texttt{orig\_group}, as we are creating a system to aid scholars who want a reconstruction of the original text of the manuscripts. Within the digitized original text, damaged and missing sections of the manuscripts are represented with brackets and dots, which can be used to convey different levels of damage and manual reconstruction in the manuscript. This information is represented in the Leiden+\footnote{\url{https://papyri.info/docs/leiden_plus}} documentation format: missing sections are denoted with brackets with dots inside, where the number of dots is equal to the estimated number of characters missing in the line of text (so [...] would indicate 3 missing characters); brackets with letters inside indicate a damaged section which was reconstructed by a scholar; and characters with some damage that have been manually reconstructed by a scholar can appear outside of brackets with a dot beneath them. Immediately below are example sentences from the data showing these formatting styles:

\begin{enumerate}[itemsep=-0.12cm]
    \item[] Blank Lacunae:
    \item[] \begin{coptic} afbeebe\textbf{[...]}af;ice\textbf{[...]}\end{coptic}
    \item[]Reconstructed Lacunae:
    \item[] \begin{coptic}auwafylhl[\textbf{af}]tnnoouf\end{coptic}
    \item[] \begin{coptic}auwmn\textbf{\stackunder[-1.5pt]{p}{.}}etc/ouorte\textbf{\stackunder[-1.5pt]{p}{.}}etconcmpnoute\end{coptic}
\end{enumerate}

The completely blank sections are the target use case for our system, and we use the manually reconstructed lacunae as the gold standard test data for our model. As this gold standard test data is a limited proportion of the corpora, we also mask characters from the sentences of the corpora without lacunae to create training data and additional test data for our model. The Coptic SCRIPTORIUM Corpora have a total of 36,252 complete sentences (no lacunae) with over 2.8 million characters. The lengths of these sentences range from 5 characters to 1067 characters, with an average sentence length of 80 characters. We created a train/dev/test data partition from these complete sentences, with the proportions 90:5:5, giving us a training data set of 32,676 sentences, a dev data set of 1,815 sentences, and a test set of 1,816 sentences. 

In addition to the complete sentences, there is a portion of sentences in the Coptic SCRIPTORIUM Corpora which contain lacunae. There are a total of 792 sentences, with approximately 60,000 characters, which contain only those lacunae that have been manually reconstructed by Coptic language scholars. This set of sentences is our gold standard test data. The average sentence length in this set is 75 characters. The total number of missing characters in this test set is 3,594, with an average gap length of \textasciitilde2 characters. There are also 780 sentences, with approximately 52,000 characters, that contain at least one empty lacuna which has not been reconstructed by a scholar. This set of sentences is the target data that we are building our system to fill in, so there is no gold standard to evaluate against directly. The average sentence length in this set is 68 characters, and the total number of missing characters is 3,658, with an average gap length of \textasciitilde3 characters. The similarities in average sentence length and average lacuna gap length between these two data sets suggest that the model should be able to perform well on the target data set if it performs well on the test data set. 

\section{Model Architecture}

\begin{figure}
  \centering
  \includegraphics[width=1\linewidth]{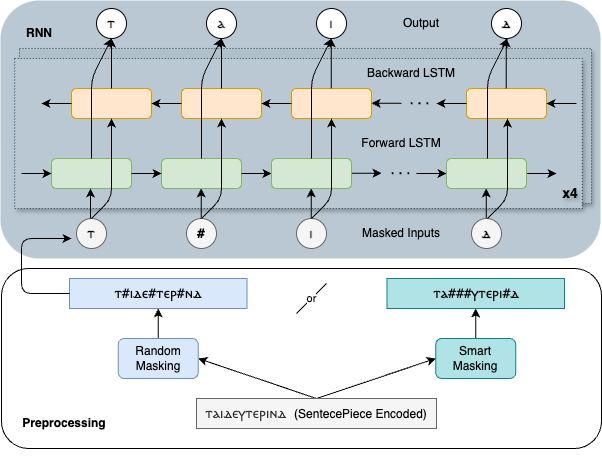}
  \caption{Model architecture and preprocessing}
  \label{fig:RNN}
\end{figure}

For our lacuna prediction model, we implement a character based bidirectional RNN model, trained with a character-level masked language modeling task. We start with a character-level vocabulary and embedding layer, generated with SentencePiece \cite{kudo-richardson-2018-sentencepiece}. The vocabulary is 134 characters, including some control symbols, the mask token, the lower-cased Coptic alphabet, and some punctuation. Our final model has an embedding size of 200, hidden size of 300, and projection size of 150. For the body of the model, we then create a four layer bidirectional LSTM, an AdamW optimizer, and learning rate of 0.0003 (selected as the optimal parameters after conducting an extensive hyperparameter search). The LSTM architecture was chosen over other architectures, such as GRU, for its ability to capture long distance dependencies, which provide relevant context for lacuna reconstruction. We use categorical cross entropy for our loss criterion, evaluating only on predictions for masked characters. A diagram of our model architecture and preprocessing is shown in Figure \ref{fig:RNN}.

The resulting model is fairly small model, and training on the full training data set can be accomplished in a few hours, with or without GPU hardware. As such, with our model code and the publicly available data from the Coptic SCRIPTORIUM project, models at the performance level presented in this paper will be accessible to interested parties. The code for our model is available on GitHub\footnote{\url{https://github.com/lauren-lizzy-levine/coptic_char_generator.git}}, and instructions for recreating the data partition, training the model, and running/interacting with the model are included in the README.

We explore several different masking strategies in the training of our model. For the first masking strategy, which we refer to as "random masking", we used the BERT masking strategy of randomly masking 15\% of the characters. When creating the index vector for the sentence, each character has a 15\% chance of being masked. If the character is masked, there are three possible masking options. 80\% of the time, the character is replaced with a special mask token, while 10\% of the time it is replaced with a random character, and finally, 10\%  of the time, the character is not replaced.

We also implement a masking strategy called "smart masking", which mimics the distribution of lacunae in the gold standard test set (described in Section \ref{data}). In the reconstructed lacuna test set, the sentences range from having one gap to as many as twenty. Over 60\% of the sentences have just one gap, 35\% have two to nine gaps, and just 5\% have more than nine gaps. To mimic the variable number of gaps, we randomly incorporate one to five gaps per sentence. Of the 1,470 gaps in the 782 sentences, almost half of them are just one character long. The length of each gap has 48\% of being just one character long, 22\% chance of being two characters long, 12\% chance of being 3 characters long, and for the final 18\% of the time, the gap length is randomly generated to be between four and thirty-four characters.

In addition to the two different masking strategies for distribution, we also had two strategies relating to the re-masking frequency of the data. The first strategy is to mask one time, when loading the data initially, which we call "once masking". The second option is to re-mask the training data at each epoch, which we call "dynamic masking". Between the two masking distribution strategies and these two re-masking frequency strategies, we ended up with four different model types: random-once, random-dynamic, smart-once, and smart-dynamic. For training, we auto-generated masked dev data that matched the distribution masking strategy (random or smart) of the model being trained.

\section{Evaluation}
\label{evaluation}

For evaluation, we had three different test sets. From the test partition made from the complete sentences that had no lacunae, we created two test sets: one with random masking and one with smart masking. Our final test data set was the gold standard data of manually reconstructed lacunae described in Section \ref{data}. We had our models predict on the data in all three test sets and scored their performance with a simple accuracy metric (number of masked characters correctly predicted / total number of masked characters in data set). 

\subsection{Baselines}
\label{baseline-results}
We applied three rudimentary heuristic baselines to our three test data sets, the results of which are shown in the bottom half of Table~\ref{tab:accuracy}. The first baseline selected a random character from the SentencePiece character model vocabulary for each character prediction. The second baseline always predicted the most common letter in the data set (mode character), "\begin{coptic}e\end{coptic}". The third baseline is a simple tri-gram language model. Results for "Test Random" and "Test Smart" are the performance of the baselines on the auto-generated random masked test data and smart masked test data respectively, while "Test Reconstructed Lacunae" is the performance on the gold standard data of manually reconstructed lacunae. 

\begin{table*}
\centering
\begin{tabular}{lccc}
\hline 
 & \textbf{Test Random} & \textbf{Test Smart} & \textbf{Test Reconstructed Lacunae } \\ \hline
\textbf{Models} &  &  &  \\
Random-Once & 0.703 & 0.323 & 0.336 \\
Random-Dynamic & \textbf{0.722} & 0.338 & \textbf{0.369} \\
Smart-Once & 0.610 & \textbf{0.366} & 0.334 \\
Smart-Dynamic & 0.603 & 0.359 & 0.319 \\
\hline
\textbf{Baselines} & & & \\
Tri-gram & 0.259 & 0.134 & 0.155 \\
Mode Character & 0.126 & 0.124 & 0.121 \\
Random & 0.008 & 0.007 & 0.007\\
\hline
\end{tabular}
\caption{Model and baseline accuracy results on the three test sets}
\label{tab:accuracy}
\end{table*}

\subsection{RNN Evaluation}
\label{sec:RNN_eval}

We started our model training by doing hyperparameter searches on four different model configurations, using combinations of the masking strategies for masking distribution and re-masking frequency (random-once model, random-dynamic model, smart-once model, smart-dynamic model). After we selected the best performing hyperparameters for each masking configuration with regard to accuracy scoring on the correspondingly masked dev data, we ran the four best performing models (one for each masking configuration) on the three test data sets outlined at the top of this section. The results from these runs are shown in the top section of Table \ref{tab:accuracy}.

The random test set has the highest scores on average, while the reconstructed lacuna test set has the lowest scores on average, indicating that the reconstructed lacuna test set is the more difficult scenario. However, it is also the most realistic scenario out of all three test sets, so performance on this test set should be considered the most significant. We observe that all of the tested model configurations outperform the baselines, showing a substantial increase in performance on all test sets. Out of the four different masking strategies we explored, we found that the model utilizing the random-dynamic masking strategy had the highest performance on the random test set and the reconstructed lacuna test set, while the smart-once masking strategy had the highest performance on the smart test set.

\begin{figure}
  \centering
  \includegraphics[width=1\linewidth]{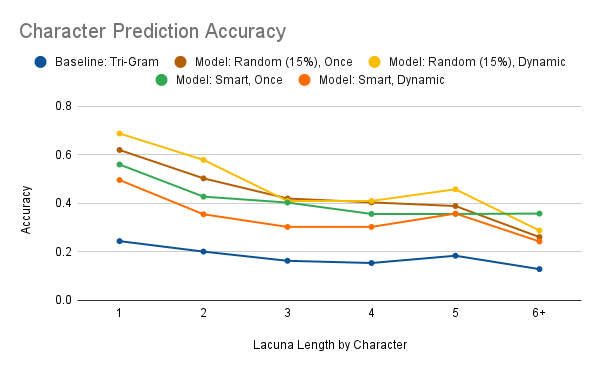}
  \caption{Accuracy of the various model configurations and tri-gram baseline relative to lacuna length in characters}
  \label{fig:accuracy}
\end{figure}

It is somewhat surprising that the model utilizing the random strategy outperforms the model using the smart strategy on the reconstructed lacuna set, considering that the smart masking strategy was developed to better reflect the conditions in which actual lacunae occur. This result is likely because the reconstructed lacuna data set is composed of only sentences with fully reconstructed lacunae, and thus is biased towards containing shorter lacunae than we might otherwise expect. As such, in Figure \ref{fig:accuracy} we consider the accuracy of each of our models with respect to the length (in characters) of the lacuna being reconstructed, and we observe that overall performance decreases as lacuna length increases. We also see that while the random-dynamic model has the best performance for lacunae of length 1-2, the smart-once model actually has better performance for lacunae of length 6+. For this reason, we recommend the smart-once model configuration for cases where the lacuna is more than a few characters. For our use case studies in Section \ref{sec:case_studies}, we consider outputs from both the random-dynamic model and the smart-once model. 

\subsection{Relative Ranking}

As we saw in the previous section evaluating the quality of our RNN model outputs, performance on the more realistic reconstructed lacuna test set was relatively low, peaking at 37\% accuracy. As such, we cannot consider the model by itself to be a definitive means of manuscript reconstruction. The model is better thought of as an additional tool in the toolbox of scholars attempting to reconstruct manuscript lacunae. To this end, we propose to use the RNN model as a means of ranking the likelihood of potential candidates for the lacuna reconstruction. 

If a scholar has several candidates for a lacuna from various qualitative methods of reconstruction, in addition to getting the model to predict what it considers to be the most likely reconstruction, we can also extract the probabilities associated with each of the scholar's potential reconstructions. Once we have these probabilities, we can sort them in descending order to get a ranking of which potential reconstructions the model considers to be more likely.\footnote{One limitation of this is that in order for the probabilities to be compatible, the input context for the model must be the same. This means that all candidates being compared must be of the same character length.} This will give a sense of which option is statistically the most likely considering the distribution of characters present in the training data of the model. While still not definitive, this ranking gives scholars another piece of evidence to consider when putting forward an argument for a particular reconstruction. 

\section{Case Studies}
\label{sec:case_studies}

In this section, we demonstrate how our RNN model may be integrated into the workflow of a Coptic scholar working on manuscript reconstruction by looking at use cases in two early Coptic manuscripts. We use the model to predict reconstructions, or to produce relative rankings of potential reconstructions under consideration. We explore how this additional information may contribute to a scholar's considerations during the reconstruction process.

\subsection{Isaiah 37:24}
\label{sec:isaiah}

\begin{figure}
  \centering
  \includegraphics[width=0.75\linewidth]{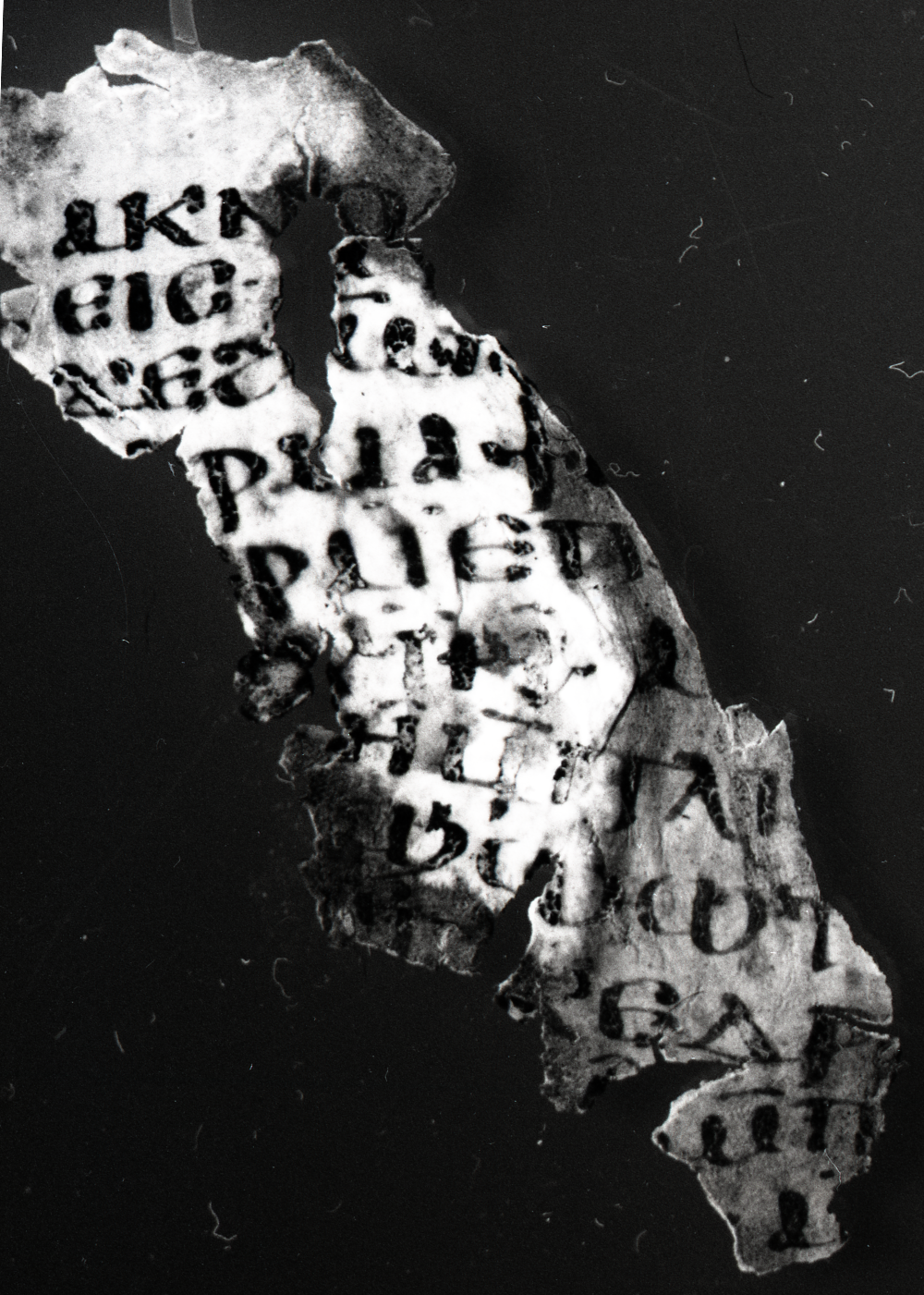}
  \caption{P.Duk. inv. 282 fr. B verso}
  \label{fig:img048}
\end{figure}

The manuscript P.Duk. inv. 282 comprises four contiguous fragments from a parchment codex and is currently held at Duke University, pictured in Figure \ref{fig:img048} \cite{Wagner2022}. The manuscript contains portions of Isaiah chapters 36--38 in the Sahidic dialect. The manuscript's date is unknown. Some lacunae in the manuscript can be restored on the basis of the only other Sahidic manuscript containing these chapters, Morgan Library M 568. For example, at Isaiah 36:16 there are two small lacunae in the Duke manuscript: \begin{coptic}a[..]n\stackunder[-1.5pt]{t}{.}e[.]\=n\-cemoou\end{coptic}. We can restore the original reading with confidence from the Morgan manuscript: \begin{coptic}a[uw] n\stackunder[-1.5pt]{t}{.}e[t]\=n\-ce moou\end{coptic} (``and you will drink water''). 

Other lacunae in the Duke manuscript cannot be restored entirely on the basis of the Morgan manuscript. For example, at Isaiah 37:24 there are four lacunae: \begin{coptic}akno[.......]eic\0a[.....]djeh1m\stackunder[-1.5pt]{\=p}{.}[.......]a\stackunder[-1.5pt]{r}{.}matj[.......]ra\"iep\end{coptic}\\\begin{coptic}[....]\end{coptic}. The Morgan manuscript helps restore the passage excluding the penultimate lacuna: \begin{coptic}akno[qneq pdjo]eic a[kdjooc] djeh1m\stackunder[-1.5pt]{\=p}{.}[aya\"i nnh1]a\stackunder[-1.5pt]{r}{.}ma tj[..... eh1]ra\"i ep[djice]\end{coptic} (``You have reproached the Lord. You said, `with the multitude of my chariots [...] to the height.''). Where the Duke manuscript has \begin{coptic}tj[\end{coptic}, the Morgan manuscript contains the past tense conjugation base verb: \begin{coptic}a\"iale eh1ra\"i epdjice\end{coptic} (``I have gone up to the height.''). Still, as in the Morgan manuscript, the Duke manuscript must contain a verb in the lacuna followed by \begin{coptic}eh1rai\end{coptic} ``up''. Thus the letter before the lacuna can only be the personal subject prefix \begin{coptic}tj\end{coptic} - ``I'', which must be followed by a present or future tense verb. Restricting our search to verbs that are both contextually appropriate and appear in high frequency in the database of the Coptic SCRIPTORIUM project, we propose three reconstructions. The models rank each reconstruction as a sequence of consecutive characters, including uninterrupted context following and especially before the gap: \begin{coptic}aknoqneqpdjoeicakdjoocdjeh1mpayainnh1armatj[.....]eh1rai\end{coptic}. The three reconstructions from the random-dynamic model are as follows, in order of probability (the log probability\footnote{Log probabilities are used to avoid potential numerical underflow that can result from the multiplication of standard probabilities when calculating the likelihood of a sequence.} of each sequence is included in parenthesis):

\vspace{12pt}
\begin{enumerate}
     \item \begin{coptic}mooye\end{coptic} (-11.16) $\rightarrow$ \begin{coptic}tj[mooye] eh1rai\end{coptic} \\``I am walking up''
   \item \begin{coptic}nabwk\end{coptic} (-12.27) $\rightarrow$ \begin{coptic}tj[nabwk] eh1rai\end{coptic} \\``I will go up”
    \item \begin{coptic}naale\end{coptic} (-12.60) $\rightarrow$ \begin{coptic}tj[naale] eh1rai\end{coptic} \\``I will rise up''
\end{enumerate}

The first result, which is in the durative present tense, is less appealing than the other results when considering the other ancient language witnesses to this passage in Isaiah. To the best of our knowledge, all witnesses approximate the Morgan manuscript’s past tense, except for two witnesses that give the future: the Syriac Peshitta (\textit{’n’ ’sq}, `I will go up'') and some manuscripts of Jerome’s \textit{Commentary on Isaiah} (11.7: \textit{ego ascendam}, ``I will go up''). These two witnesses increase the probability of the second and third result, both of which are in the future tense (signaled by the auxiliary \begin{coptic}na\end{coptic}). Although ranked lower by the model, some scholars would surely prefer the third result over the second, since, as we saw above, the same verb (\begin{coptic}ale\end{coptic}) appears in this passage in the Morgan manuscript. On the other hand, beyond its higher ranking, \begin{coptic}bwk\end{coptic} appears far more often in Old Testament books and especially in Isaiah: in the Morgan manuscript \begin{coptic}bwk\end{coptic} is used five times in ch. 37 alone, while \begin{coptic}ale\end{coptic} appears only here at the point of disagreement with the Duke manuscript. Thus the most plausible restoration of the passage: \begin{coptic}a[kdjooc] djeh1m\stackunder[-1.5pt]{\=p}{.}[aya\"i nnh1]arma tj[nabwk eh1]ra\"i epdjice\end{coptic} ``You have reproached the Lord. You said, `with the multitude of my chariots I will go up to the height.''

\subsection{The Nag Hammadi Library -- \textit{Gospel of Philip}}

The \textit{Gospel of Philip} (\textit{GPhilip}) is the third composition included in codex II of the Nag Hammadi (NH) library, a collection of thirteen papyrus codices containing a diverse range of ancient Christian texts. Unlike the example discussed in Section \ref{sec:isaiah}, there are no other surviving manuscript versions of (\textit{GPhilip}). The codex sustained moderate damage to the top and bottom margins and most of its leaves contain peninsula-shaped lacunae\footnote{Archival photo of the manuscript: \url{https://ccdl.claremont.edu/digital/collection/nha/id/2962/rec/182}}.

The restoration of Saying 55 (63.30-64.5) has been a particular point of scholarly intrigue. While smaller gaps in the Saying can be restored with some confidence, scholars have proposed various readings for one lacuna of 5-6 letters, which contains the object of the verb \begin{coptic}acpaze\end{coptic} or ``kiss.'' The passage, which describes Jesus kissing Mary Magdalene, reads: \begin{coptic}nef1acpaze mmoc atec...nh1ah1 ncop\end{coptic}, "He used to kiss her on the ..... many times" (63.35-36). 

This case presents an especially challenging reconstruction due to the size of the lacuna. As discussed above (Section \ref{sec:RNN_eval}), the accuracy of the model degrades as the size of the lacuna increases. We consider outputs from both the smart-once model, which provides the highest accuracy rates for longer lacunae, and the random-dynamic model, which provides the highest accuracy rates for short lacunae.   

Since the model is trained on Sahidic texts, the Saying needs to follow the orthographic conventions of the Sahidic dialect. Thus we changed the prenominal preposition \begin{coptic}a\end{coptic}- `towards, on' to \begin{coptic}e\end{coptic}- (in the Sahidic dialect \begin{coptic}a\end{coptic} is the past tense marker), resulting in the input text: \begin{coptic}acpaze \"mmoc etec[.....] \"nh1ah1 \"ncop\end{coptic} ``kissed her on her ... on many occasions''. 

The four letters before the lacuna includes an indirect object construction headed with preposition \begin{coptic}a\end{coptic} followed by a feminine possessive article \begin{coptic}tec\end{coptic}- ``hers.'' Due to this syntactic environment, reconstructions are limited to feminine nouns, likely a body part in this case. To fill the lacuna, we have the models produce their predictions for either a 5 character gap or a 6 character gap: 

\begin{enumerate}[itemsep=-0.15cm]
    \item[] Smart-Once:
    \begin{enumerate}[itemsep=-0.05cm]
        \item[] 5 spaces: \begin{coptic}h1huen\end{coptic}
        \item[] 6 spaces: \begin{coptic}h1hueee\end{coptic}
    \end{enumerate}
    \item[] \vspace{6pt} Random-Dynamic:
    \begin{enumerate}[itemsep=-0.05cm]
        \item[] 5 spaces: \begin{coptic}h1ooee\end{coptic}
        \item[] 6 spaces: \begin{coptic}h1ooeee\end{coptic}
    \end{enumerate}
\end{enumerate}
Unfortunately, none of the reconstructions produce an attested Coptic lemma. 

However, the models can still be leveraged to compare editorial suggestions and assign greater or lesser probability of editorial reconstructions. In this case, editors Bentley Layton and Hans-Martin Schenke propose several  options for a 5 letter feminine body part: Schenke proposes ``mouth'' (\begin{coptic}tapro\end{coptic}). Layton offers multiple readings: ``mouth'' (\begin{coptic}paiqe\end{coptic} or \begin{coptic}tapro\end{coptic}), ``cheek'' (\begin{coptic}ouoqe\end{coptic}), ``foot'' (\begin{coptic}qalodj\end{coptic}), and "forehead"(\begin{coptic}teh1ne\end{coptic}) as possible candidates \cite{Schenke1997,LaytonIsenberg1989}. The editors present these candidates in an unordered manner, not singling out any one as being particularly more likely than the others. 

\begin{table*}
\centering
\begin{tabular}{cccc}
\hline 
\textbf{Smart-Once Norm} & \textbf{Smart-Once Orig} & \textbf{Random-Dynamic Norm} & \textbf{Random-Dynamic Orig} \\ 
\hline
\begin{tabular}[c]{@{}c@{}}\begin{coptic}h1huen\end{coptic} \textit{NA} \\ (-6.89)\end{tabular}  & 
\begin{tabular}[c]{@{}c@{}}\begin{coptic}h1huen\end{coptic} \textit{NA} \\(-7.69)\end{tabular} & 
\begin{tabular}[c]{@{}c@{}}\begin{coptic}h1ooee\end{coptic} \textit{NA} \\(-7.88) \end{tabular} & 
\begin{tabular}[c]{@{}c@{}}\begin{coptic}h1ooee\end{coptic} \textit{NA} \\(-8.05) \end{tabular} \\
\hline
\begin{tabular}[c]{@{}c@{}}\begin{coptic}h1ooee\end{coptic} \textit{NA} \\ (-8.08) \end{tabular} & 
\begin{tabular}[c]{@{}c@{}}\begin{coptic}h1ooee\end{coptic} \textit{NA} \\ (-7.99)  \end{tabular}& 
\begin{tabular}[c]{@{}c@{}}\begin{coptic}h1huen\end{coptic} \textit{NA} \\ (-11.51) \end{tabular} & 
\begin{tabular}[c]{@{}c@{}}\begin{coptic}h1huen\end{coptic} \textit{NA} \\ (-11.90) \end{tabular}\\
\hline
\begin{tabular}[c]{@{}c@{}}\begin{coptic}ouoqe\end{coptic} \textit{cheek} \\ (-16.11) \end{tabular} & 
\begin{tabular}[c]{@{}c@{}}\begin{coptic}ouoqe\end{coptic} \textit{cheek} \\(-15.64) \end{tabular} & 
\begin{tabular}[c]{@{}c@{}}\begin{coptic}teh1ne\end{coptic} \textit{forehead} \\ (-12.95) \end{tabular} & 
\begin{tabular}[c]{@{}c@{}}\begin{coptic}teh1ne\end{coptic} \textit{forehead} \\ (-13.08) \end{tabular}\\
\hline
\begin{tabular}[c]{@{}c@{}}\begin{coptic}teh1ne\end{coptic} \textit{forehead} \\ (-16.53) \end{tabular} &
\begin{tabular}[c]{@{}c@{}}\begin{coptic}teh1ne\end{coptic} \textit{forehead} \\ (-16.42) \end{tabular} &
\begin{tabular}[c]{@{}c@{}}\begin{coptic}ouoqe\end{coptic} \textit{cheek} \\ (-13.16) \end{tabular} &
\begin{tabular}[c]{@{}c@{}}\begin{coptic}ouoqe\end{coptic} \textit{cheek} \\ (-14.39) \end{tabular}\\
\hline
\begin{tabular}[c]{@{}c@{}}\begin{coptic}qalodj\end{coptic} \textit{foot} \\ (-17.35) \end{tabular} &
\begin{tabular}[c]{@{}c@{}}\begin{coptic}qalodj\end{coptic} \textit{foot} \\ (-17.42) \end{tabular} &
\begin{tabular}[c]{@{}c@{}}\begin{coptic}tapro\end{coptic} \textit{mouth} \\ (-14.66)  \end{tabular} &
\begin{tabular}[c]{@{}c@{}}\begin{coptic}tapro\end{coptic} \textit{mouth} \\ (-14.79) \end{tabular}\\
\hline
\begin{tabular}[c]{@{}c@{}}\begin{coptic}paiqe\end{coptic} \textit{mouth}\\ (-18.74) \end{tabular} & 
\begin{tabular}[c]{@{}c@{}}\begin{coptic}tapro\end{coptic} \textit{mouth} \\ (-18.64) \end{tabular} & 
\begin{tabular}[c]{@{}c@{}}\begin{coptic}paiqe\end{coptic} \textit{mouth} \\ (-16.12) \end{tabular} & 
\begin{tabular}[c]{@{}c@{}}\begin{coptic}paiqe\end{coptic} \textit{mouth} \\ (-15.36) \end{tabular}\\
\hline
\begin{tabular}[c]{@{}c@{}}\begin{coptic}tapro\end{coptic} \textit{mouth} \\ (-19.02) \end{tabular}& 
\begin{tabular}[c]{@{}c@{}}\begin{coptic}paiqe\end{coptic} \textit{mouth} \\ (-18.71) \end{tabular}& 
\begin{tabular}[c]{@{}c@{}}\begin{coptic}qalodj\end{coptic} \textit{foot} \\ (-16.94) \end{tabular}& \begin{tabular}[c]{@{}c@{}}\begin{coptic}qalodj\end{coptic} \textit{foot} \\ (-16.48) \end{tabular}\\
\end{tabular}
\caption{Rankings of lacuna candidates for the \textit{GPhilip} use case (English translation in italics and log probabilities in parenthesis)}
\label{tab:rankings}
\end{table*} 

Table \ref{tab:rankings} compares the output of smart-once model and the random-dynamic model, and contrasts the confidence of each model's top two predictions (again, not attested Coptic lemmas) with the list of attested feminine nouns supplied by the editors. As the table details, the lemma (\begin{coptic}ouoqe\end{coptic}), ``cheek'' is favored by the smart-once model and the lemma (\begin{coptic}teh1ne\end{coptic}), ``forehead'' is preferred by the random-dynamic model. Both of these results differ from Schenke's reconstruction of \begin{coptic}tapro\end{coptic}, ``mouth'' \cite{Schenke1997}. 

Table \ref{tab:rankings} also compares the effect of normalization on the model reconstructions. As discussed above in Section \ref{data}, the Coptic SCRIPTORIUM data utilized to train the model includes both normalized and original (un-normalized) data. We hypothesized that the normalization of dialect differences to conform to Sahidic orthography would greatly impact the results. However, in the end, the normalization had little impact and only slightly modified the ranking orders and confidence as Table \ref{tab:rankings} documents. Note the slightly different ranking of \begin{coptic}paiqe\end{coptic} and \begin{coptic}tapro\end{coptic}, two different words meaning "mouth," by the smart-once model. 

These models provide quantitative data about reconstructions and offers a relative ranking of the alternatives proposed by text editors. In cases like the one discussed above in \textit{GPhilip} where editors have provided multiple possible reconstructions to fill the lacuna and comparison to other manuscripts is not possible, this is an especially valuable tool in assisting readers in deciding which reading best fits within their comprehension of the passage. 

\section{Conclusion}

In this paper, we presented a bidirectional RNN architecture to reconstruct lacunae in Coptic manuscripts. When training our masked language model for character prediction, we explored different masking strategies  for masking distribution (random and smart) and re-masking frequency. We evaluated our models against both artificially masked data and scholar-reconstructed lacunae. We found that the performance of our models declined as the length of the lacunae being reconstructed increased, peaking at above 70\% for single character reconstruction and below 40\% for lacunae of length 6+ characters. And while the model trained with random masking performed with higher accuracy for single character reconstruction, the model trained with smart masking performed with higher accuracy on the reconstruction of longer lacunae, which is more similar to the real world use case, as it is more difficult for scholars to qualitatively reconstruct longer lacunae.

Using the judgments from these models, we explored two use cases of lacuna reconstruction from ancient Coptic manuscripts. We considered not only the direct predictions from the models, but also the likelihood ranking of reconstruction candidates already under consideration from the past proposals of various scholars of Coptic. Despite the low accuracy of the models on reconstructing lacunae of more than a few characters, we see that the rankings can still be leveraged to provide additional quantitative evidence alongside traditional qualitative methods. This initial application of neural methods to Coptic manuscript reconstruction shows the potential for integrating the judgments of models with the existing qualitative methods used by scholars working on manuscript reconstruction.

\section*{Limitations}

As previously discussed in Section \ref{evaluation}, the quality of our RNN models is relatively low, limiting the utility of its judgments. As we primarily consider a single model architecture in this investigation, in future work it would be beneficial to explore architectures beyond RNNs and training tasks beyond masked language modeling. In addition to different architectures, we believe there is much room for exploring different inputs for model training, including lexicographic information (what possible words might be, for example using a digital Coptic dictionary such as \citealt{feder-etal-2018-linked}), or linguistic annotations, such as morphosyntactic information provided by Coptic treebank data and corresponding parsers \cite{zeldes-abrams-2018-coptic,zeldes-schroeder-2016-nlp}. 

Additionally, our current model does not account for the diacritics used in Coptic writing, and it is trained on a sentence-wise basis without incorporating document-level information, such as the surrounding sentences, or details about the page layout. Future work may benefit from incorporating diacritics and additional context into the training paradigm for the model. Future work should also include the ability to give a ranking of lacuna candidates of different lengths, which is not currently possible because model inputs must be of the same sequence length for their probabilities to be comparable.

\section*{Acknowledgements}

This research was generously supported by a major grant from the National Endowment for the Humanities Humanities Collections and Reference Resources (PW-290519-23). 

\bibliography{custom}

\appendix



\end{document}